\documentclass{article} 
\usepackage{xgenact_preprint,times}

\usepackage{amsmath,amsfonts,bm}

\def\eqref#1{equation~\ref{#1}}

\def\1{\bm{1}}

\def\vtheta{{\bm{\theta}}}

\def\ve{{\bm{e}}}

\def\vn{{\bm{n}}}

\def\vp{{\bm{p}}}
\def\vq{{\bm{q}}}

\def\vx{{\bm{x}}}

\def\mR{{\bm{R}}}

\DeclareMathAlphabet{\mathsfit}{\encodingdefault}{\sfdefault}{m}{sl}
\SetMathAlphabet{\mathsfit}{bold}{\encodingdefault}{\sfdefault}{bx}{n}
\newcommand{\tens}[1]{\bm{\mathsfit{#1}}}

\def\tC{{\tens{C}}}

\def\gD{{\mathcal{D}}}
\def\gE{{\mathcal{E}}}

\newcommand{\etens}[1]{\mathsfit{#1}}

\def\etA{{\etens{A}}}

\newcommand{\R}{\mathbb{R}}

\usepackage[hidelinks]{hyperref}
\usepackage{url}

\usepackage{tikz}
\usepackage{adjustbox}
\usetikzlibrary{positioning}
\usepackage{booktabs}
\usepackage{amssymb}           
\usepackage{tabularx}
\usepackage{multirow}
\usepackage{threeparttable}
\usepackage{graphicx}
\usepackage{wrapfig}
\usepackage{needspace}
\usepackage{xcolor}
\usepackage{colortbl}
\definecolor{tablegray}{gray}{0.94}

\newcommand{\method}{\textsc{XGenAct}}

\newcommand{\tmpl}[1]{\texttt{#1}}            
\newcommand{\menu}[1]{{\fontfamily{lmss}\selectfont #1}}  
\newcommand{\ptag}[1]{\texttt{<#1>}}          
\newcommand{\tabref}[1]{Table~\ref{#1}}
\newcommand{\appref}[1]{Appendix~\ref{#1}}

\newcommand{\ai}{\textsc{ActionImages}}      

\title{
XGenAct: Geometry-Enhanced World Action Models through Cross-Task Generation
}

\author{Tingting Du\thanks{\texttt{tdu35@wisc.edu}}\textsuperscript{1},
Ziyao Wang\textsuperscript{2}, Guoheng Sun\textsuperscript{2},
Ang Li\thanks{Correspondence to: Ang Li (\texttt{angliece@umd.edu}).}\textsuperscript{2}\\
\textsuperscript{1}University of Wisconsin--Madison\\
\textsuperscript{2}University of Maryland, College Park}

\definecolor{givenfill}{RGB}{197,218,238}
\definecolor{predfill}{RGB}{252,221,188}

\newcommand{\fitwidth}[1]{%
  \resizebox{\ifdim\width>\textwidth \textwidth\else \width\fi}{!}{#1}}

\begin{document}

\maketitle

\begin{figure}[h]
  \centering
  \includegraphics[width=\linewidth]{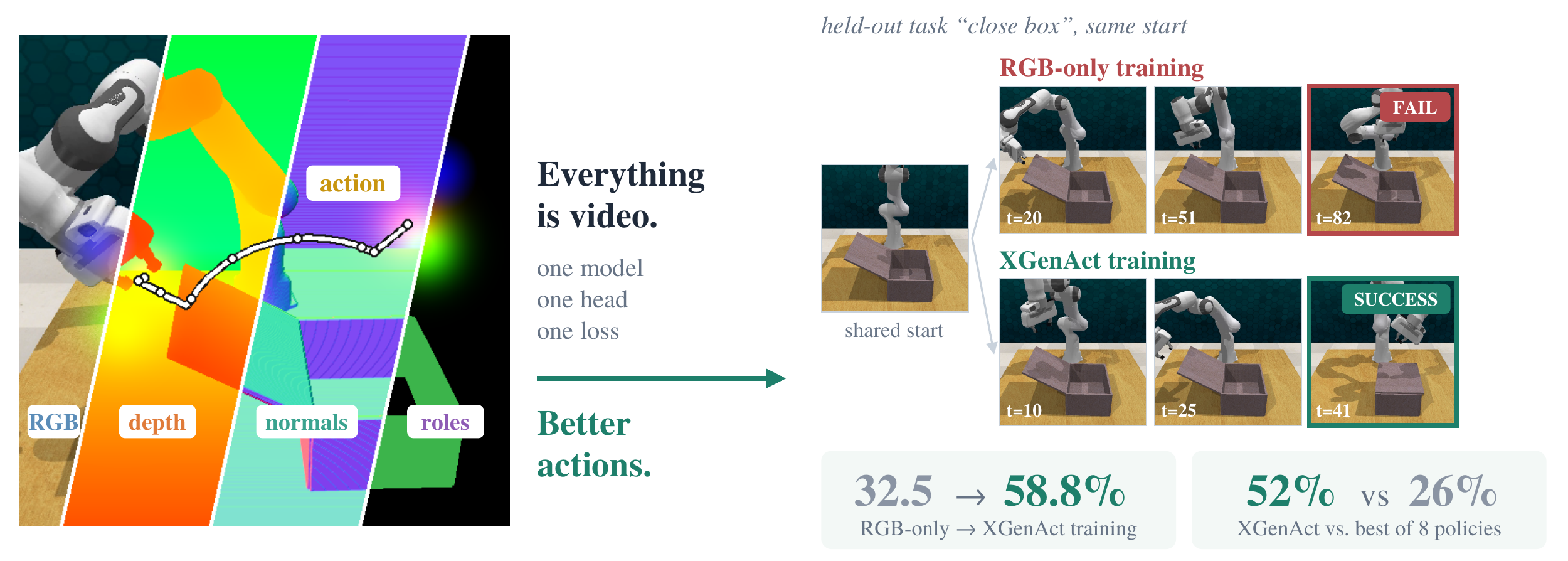}
  \caption{\textbf{One checkpoint, multiple future spaces.}
  Given initial observation and action anchors and the instruction ``slide the bottom drawer
  open,'' \method{} generates future RGB, action images, metric depth, functional-role
  segmentation, and surface normals at matched temporal indices. The $t{=}0$ column
  is given; the rest are generated. The visual rows are obtained through
  separate prompts to the same model rather than one five-stream decode; one camera view
  is shown.}
  \label{fig:teaser}
\end{figure}

\begin{abstract}
World action models (WAMs) have advanced robot control by predicting how observations and actions evolve over time. Despite this progress, RGB and action based future prediction does not explicitly address the spatial understanding needed for robot manipulation. Existing efforts often add a limited set of spatial prediction tasks through specialized heads or branches, leaving both the range of spatial supervision and the model architecture fragmented. We introduce XGenAct, a world action model that represents RGB observations, robot actions, metric depth, surface normals, and functional role segmentation as RGB videos through deterministic codecs. By sampling perception and action tasks during training, XGenAct uses one video diffusion transformer and one objective to learn temporal prediction across these spaces without modality specific learned heads. On held out RLBench tasks, structured perception training improves average closed loop success over RGB only training, and XGenAct achieves 52\% success in the five task external comparison, versus 26\% for the strongest evaluated baselines. It also predicts future depth and segmentation more accurately than the evaluated pipelines that generate RGB first and then apply a frozen perception expert.
\end{abstract}

\section{Introduction}
\label{sec:intro}

World action models (WAMs) have emerged as a promising paradigm for generative robot control by jointly predicting future RGB videos and robot actions \citep{ye2026world,zhu2025uwm,zhen2026action}. Compared with vision language action models that directly predict actions, WAMs use video generation to explicitly model how a scene evolves over time. This generative objective provides dense supervision for visual dynamics and temporal context, which can in turn improve action prediction. WAMs therefore offer a direct way to address the temporal reasoning required for robot control. However, their progress has been primarily temporal rather than spatial. Robots operate in a three dimensional world that evolves over time, while most WAMs represent the predicted world only as two dimensional RGB video. Although RGB observations contain implicit geometric cues, appearance prediction alone does not explicitly capture metric depth, surface orientation, or other spatial structures required for physical interaction. This limited spatial understanding remains a major bottleneck to further improving WAMs.

Recent work has begun to address this limitation by introducing spatial information into WAMs. One possible direction is to predict explicit three dimensional or four dimensional representations, such as point clouds or point trajectories. \citep{zhen2025tesseract, dharmarajan2025dream2flow} While these representations provide rich geometric structure, incorporating dense reconstruction directly into a generative control model requires additional representations and decoding procedures, making efficient joint prediction with robot actions nontrivial. A more practical alternative is to supervise spatial understanding through image aligned signals, including depth, optical flow, geometric features, and semantic maps. Several recent WAMs have adopted this strategy to enrich RGB based future prediction \citep{ma2026geosem,yuan2026dreamwam,li2026wam4d,zhao2026rynnworld}. However, these spatial signals are typically introduced through auxiliary prediction heads, gated residual modules, or modality specific branches, each with its own representation and training objective. Their coverage also remains fragmented. Depth receives the most attention, while few methods incorporate complementary information such as surface orientation and task dependent functional regions within a single model. We argue that spatial supervision should not remain an auxiliary component attached to an RGB based WAM. Instead, multiple geometric and semantic prediction tasks should be expressed within the same generative space used to model observations and actions, allowing structured spatial knowledge to directly shape the shared action generator.

Inspired by \ai~\cite{zhen2026action}, which represents 7 DoF robot actions as multiview RGB videos, we consider image space as a common interface for both perception and control. This representation is particularly suitable for structured visual signals because metric depth, surface normals, and segmentation can also be expressed as decodable RGB images \citep{gabeur2026image,wang2026video}. Based on this insight, we introduce \method, a geometry enhanced world action model that represents RGB observations, metric depth, surface normals, functional role segmentation, and 7 DoF robot actions as RGB videos through deterministic codecs. Every stream is encoded by the same frozen video VAE into a shared latent space. During training, each sample selects a perception action template that pairs one perception stream with the action stream, rather than combining all perception modalities simultaneously. All templates are processed by the same diffusion transformer, output head, and flow matching objective, without modality specific learned heads, auxiliary losses, or manually balanced loss coefficients. By varying the perception task across training samples, this cross task formulation exposes the shared action generator to complementary geometric and semantic supervision. Experiments on held-out RLBench tasks demonstrate that \method{} achieves the highest average closed-loop success among eight external policy baselines while retaining the ability to predict future RGB, depth, surface normals, and segmentation.

Our contributions are summarized as follows:

\begin{itemize}
\itemsep2pt

\item \textbf{Geometry enriched image space.}
We extend image space world action modeling beyond RGB and actions by encoding depth, surface normals, and functional role segmentation in the same visual modality, adding complementary geometric and semantic information.

\item \textbf{Head free and extensible task interface.}
Task templates and conditioning plans enable all perception and action tasks to share the same frozen VAE, DiT, output head, and flow matching objective, without modality specific learned heads or auxiliary losses.

\item \textbf{Improved robot control through cross-task generation.}
Experiments on held-out RLBench tasks show that structured geometric and semantic generation improves average closed-loop success by up to 28.75 percentage points over RGB-only training at matched compute and action budgets, while retaining future RGB generation.

\end{itemize}

\section{Related Work}
\label{sec:related}

\paragraph{Structured world prediction for action learning.}
World action models (WAMs) jointly learn structured visual future prediction and robot action generation, often using a shared video diffusion backbone~\citep{ye2026world,zhu2025uwm,zhen2026action,zhou2026tauzero,chen2026flowwam}. While most WAMs predict RGB futures, recent methods introduce geometric or semantic supervision. PAD~\citep{guo2024pad} extends joint image and action denoising to depth using separate modality encoders and decoders. GeoSem-WAM~\citep{ma2026geosem}, DreamWAM~\citep{yuan2026dreamwam}, WAM4D~\citep{li2026wam4d}, and X-WAM~\citep{guo2026xwam} incorporate geometry through auxiliary branches, residual modules, or training-only readouts. RynnWorld-4D~\citep{zhao2026rynnworld} generates RGB, depth, and optical flow using a three branch architecture, then predicts actions with a downstream inverse dynamics head. These methods demonstrate the value of structured future supervision, but rely on modality specific components, objectives, or action modules.

SA-WAM~\citep{lopetegui2026sawam} is closest to our representation. It encodes RGB and metric depth with the same frozen video tokenizer and jointly denoises them with actions in one latent sequence. However, actions occupy dedicated latent positions, and the model uses a fixed RGB, depth, and action layout. In contrast, \method{} represents both structured perception and action as video streams and samples multiple perception action templates within one shared generative model.

\paragraph{Perception as generation.}
A complementary line of work formulates visual understanding through image or video generation. Painter~\citep{wang2023painter} and DICEPTION~\citep{zhao2025diception} express multiple perception tasks in a common visual space. Vision Banana~\citep{gabeur2026image} and GenCeption~\citep{wang2026video} generate depth, segmentation, surface normals, and correspondence from task prompts. SUV~\citep{yuan2026suv} applies a similar formulation to driving, but retains a separate action expert. These methods establish generation as a common interface for perception, but do not examine whether a mixture of structured perception tasks can improve a robot action generator expressed in the same output space.

\paragraph{Visual representations of action.}
RT-Trajectory~\citep{gu2023rttrajectory} and TraceVLA~\citep{zheng2025tracevla} augment observations with visual motion cues but still predict action tokens. PIVOT~\citep{nasiriany2024pivot} visualizes candidate actions for a vision language model to select. Action Images~\citep{zhen2026action} instead represents a 7-DoF end effector pose as multiview RGB canvases generated by a video model. \method{} extends this image space beyond RGB observations and actions by incorporating metric depth, surface normals, and functional role segmentation. This connects perception as generation with action as generation, allowing multiple geometric and semantic tasks to supervise one action generator without modality specific heads or auxiliary losses.

\section{Method}
\label{sec:method}

\subsection{Overview}

We build \method{} on the Action Images formulation~\citep{zhen2026action} using a Wan2.2-TI2V-5B video diffusion backbone~\citep{wan2025}. Following Action Images, a 7-DoF robot action is projected into multiple camera views and rendered as an RGB video that can be decoded back into continuous control. We retain this action representation without modification. The video VAE remains frozen, and only the diffusion transformer is fine-tuned.

Our key idea is to extend this shared video space with geometry-related perception tasks. RGB observations, metric depth, surface normals, functional role segmentation, and robot actions are converted into RGB videos through deterministic codecs and encoded by the same frozen VAE. This places appearance, geometry, semantics, and control in a common latent space without introducing modality specific learned encoders or output heads.

For each training sample, we select one perception modality \(m\) and pair it with the action stream from two camera views:

\begin{equation}
\label{eq:3.1}
\Pi_m =
\left[
m^{(1)},\, a^{(1)},\, m^{(2)},\, a^{(2)}
\right],
\qquad
m\in\{\mathrm{RGB},\mathrm{depth},\mathrm{normal},\mathrm{segmentation}\}. 
\end{equation}

Thus, a sample contains one perception modality and action, rather than all perception modalities simultaneously. The template \(\Pi_m\) determines which streams the model processes, while a conditioning plan determines which frames are observed and which frames must be generated. All templates share the same diffusion transformer, output head, and training objective.

\begin{figure*}[h]
  \centering
  \includegraphics[width=\textwidth]{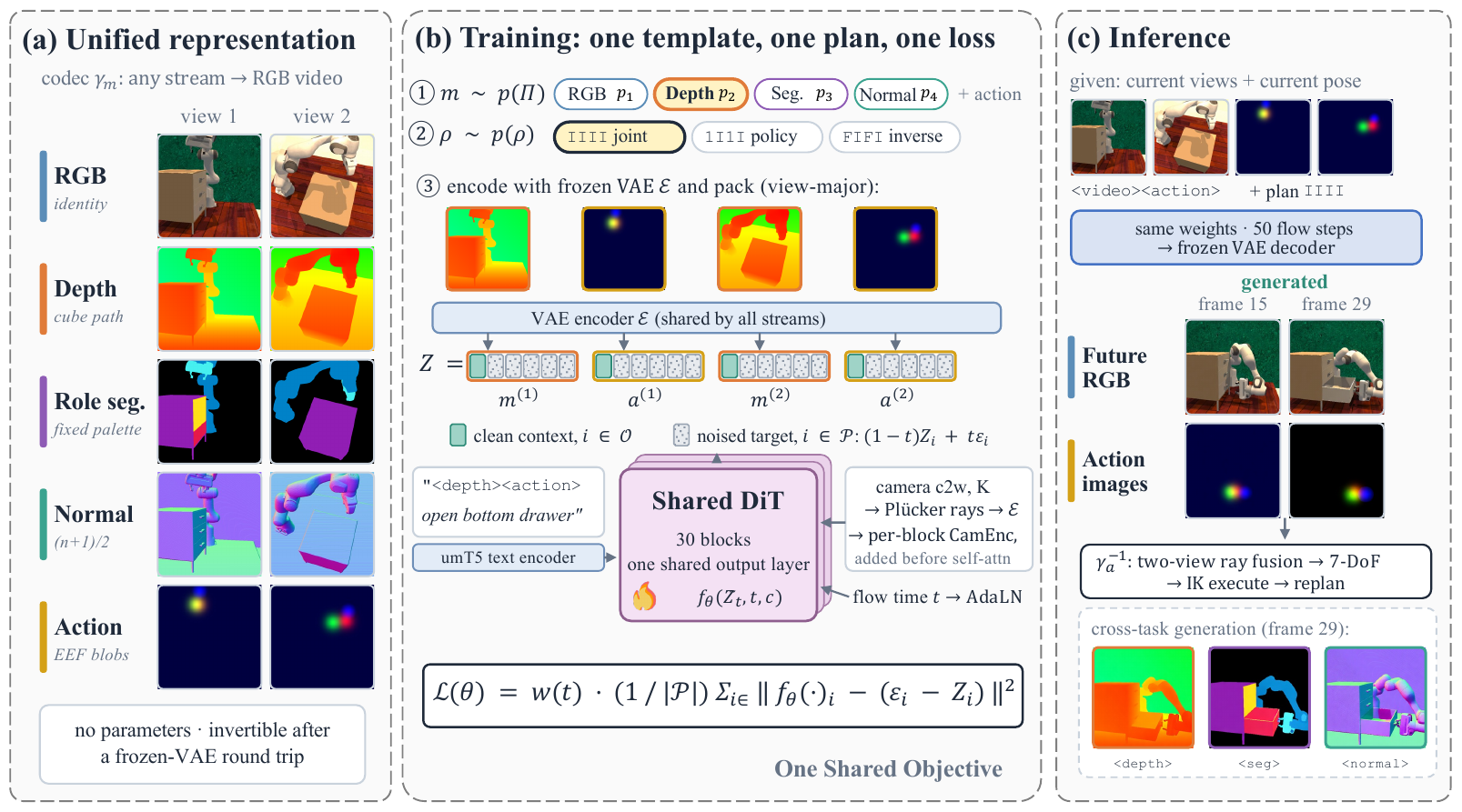}
  \caption{\textbf{Training and inference, shared architecture, and conditioning plans.}
  }
  \label{fig:method}
\end{figure*}

\subsection{Geometry Enriched Video Space}

Let \(x^{(m,v)}\) denote a video stream of modality \(m\) from camera view \(v\). We define a deterministic codec \(\gamma_m\) that converts each stream into a three channel video accepted by the frozen video VAE:

\begin{equation}
\tilde{x}^{(m,v)}=\gamma_m\!\left(x^{(m,v)}\right),
\qquad
z^{(m,v)}=\mathcal{E}\!\left(\tilde{x}^{(m,v)}\right),
\end{equation}

where \(\mathcal{E}\) is shared across all modalities. These codecs contain no learnable parameters. They preserve the information required to recover the original signal while producing videos that remain stable under VAE encoding and decoding.

\paragraph{RGB and action.}
RGB observations use the identity mapping. Robot actions use the Action Images codec~\citep{zhen2026action}, which projects each 7-DoF end effector pose into the camera views and renders the resulting trajectories as RGB videos. We use this representation unchanged.

\paragraph{Metric depth.}
Raw metric depth has an unbounded and highly nonuniform value distribution, making direct grayscale encoding inefficient. We first compress depth values with a monotonic power transform and then map the resulting scalar values onto a continuous path through the RGB cube. At decoding time, generated colors are projected back onto this path before the scalar transform is inverted. This encoding distributes metric depth across an RGB path instead of a single grayscale axis. Full definitions and decoding rules are provided in Appendix~\ref{app:codecs}.

\paragraph{Surface normals.}
A surface normal \(\mathbf{n}\in[-1,1]^3\) is mapped directly to RGB as

\begin{equation}
\gamma_{\mathrm{normal}}(\mathbf{n})
=
\frac{\mathbf{n}+\mathbf{1}}{2}.
\end{equation}

The inverse mapping recovers the components and renormalizes them to unit length.

\paragraph{Functional role segmentation.}
We assign each functional role a fixed RGB color code shared across tasks, episodes, camera views, and frames. The labels describe how a region participates in the task, such as target object, goal region, robot, tool, fixture, distractor, or background, rather than its semantic object category. This provides task relevant structural supervision while remaining compatible with the same video representation.

Before training, each codec is evaluated both directly and after a frozen VAE round trip. A new perception task can therefore be added through a deterministic codec, without introducing a learned modality specific encoder, decoder, or output head.

\subsection{Task Templates and Conditioning Plans}

Each training sample selects one perception modality \(m\) and forms the template introduced in Eq. \ref{eq:3.1}.
The four segments correspond to perception and action streams from two camera views. Their order remains fixed across modalities. Sampling different templates changes the perception task associated with action generation, while the model architecture remains unchanged.

A conditioning plan determines which parts of the packed sequence are observed. For each segment, we use one of three states: \(F\) provides the full segment, \(I\) provides its initial frame and leaves the rest to be generated, and \(1\) includes only the initial frame. This yields three settings:

\begin{equation}
\begin{aligned}
\rho_{\mathrm{joint}} &= (I,I,I,I),\\
\rho_{\mathrm{policy}} &= (1,I,1,I),\\
\rho_{\mathrm{inverse}} &= (F,I,F,I).
\end{aligned}
\end{equation}

The joint plan predicts future perception and actions together. The policy plan predicts future actions from the current perception. The inverse dynamics plan predicts actions from a fully observed perception sequence. The same plans apply to every perception modality.

During training, both the template and conditioning plan are sampled for each example. All templates use the same plan distribution. We also occasionally remove the action segments, allowing the model to learn future perception generation from an initial frame alone. Sampling probabilities are provided in the experimental setup.

\subsection{Shared Training Objective}

Let \(Z_{\Pi}\) be the packed latent sequence produced by template \(\Pi\). A conditioning plan divides its latent positions into an observed set \(\mathcal{O}\) and a prediction set \(\mathcal{P}\). Observed positions remain clean, while prediction positions are interpolated with Gaussian noise:

\begin{equation}
\tilde{Z}_{t,i} =
\begin{cases}
Z_i, & i\in\mathcal{O},\\
(1-t)Z_i+t\epsilon_i, & i\in\mathcal{P},
\end{cases}
\qquad
\epsilon\sim\mathcal{N}(0,I).
\end{equation}

The diffusion transformer predicts the flow from the clean latent to noise. We apply the flow matching loss only to positions that must be generated:

\begin{equation}
\mathcal{L}(\theta)
=
\mathbb{E}_{\Pi,\rho,t,\epsilon}
\left[
\frac{w(t)}{|\mathcal{P}|}
\sum_{i\in\mathcal{P}}
\left\|
f_{\theta}(\tilde{Z}_{t},t,\rho,\mathbf{c})_i
-
(\epsilon_i-Z_i)
\right\|_2^2
\right],
\label{eq:loss}
\end{equation}

where \(w(t)\) is the flow-scheduler weighting and \(\mathbf{c}\) contains the language instruction and camera conditioning. Observed positions provide context but do not contribute to the reconstruction loss.

This objective is identical for every perception modality, template, and conditioning plan. Cross task training changes only the content of the shared latent sequence and which positions are predicted. It introduces no modality specific output heads, auxiliary objectives, or manually chosen loss weights.

\section{Experiments}
\label{sec:experiments}


We evaluate \method{} on three questions:
(i) how it compares with existing robot policies in closed loop evaluation
(\S\ref{sec:exp_external_policies});
(ii) whether generating structured perception during training improves the action policy, and
how perception spaces affect the action performance(\S\ref{sec:exp_ablation}); and
(iii) whether generating future geometry and semantics is more accurate than generating
RGB and applying a frozen geometry perception model (\S\ref{sec:exp_main}).

\subsection{Setup}
\label{sec:exp_setup}

\paragraph{Training menus.}
We name a training configuration by the perception templates in its menu:
\menu{RGB} trains on \tmpl{video+action} only, and \menu{D}, \menu{N} and \menu{S} add
\tmpl{depth+action}, \tmpl{normal+action} and \tmpl{segmentation+action}, respectively. The most
inclusive menu, \menu{RGB+D+N+S}, draws \tmpl{video+action} with probability $0.4$ and each
perception template with probability $0.2$. Every model is fine-tuned from the released Action
Images checkpoint \citep{zhen2026action} with the same backbone, frozen VAE, objective,
conditioning-plan distribution and optimizer (effective batch 4, learning rate $5\times10^{-7}$; \appref{app:optimization}), and is
evaluated at step 4{,}000.

\paragraph{Training groups.}
We train three groups of models, one per question. Two training sources are used: a
self-generated RLBench tree \citep{james2020rlbench} of 16 tasks and $3{,}960$ episodes at
$512^2$, and a ManiSkill3 tree \citep{taomaniskill3} of 7 tasks and $1{,}750$ episodes
(\appref{app:data}).
(i) For the comparison with robot policies, \method{} (\menu{RGB+D+N+S}) and all baselines except
the zero-shot MolmoAct are fine-tuned on the same RLBench tree.
(ii) For the menu ablation, eight models, one per menu, are trained on RLBench and ManiSkill3
sampled at a $70/30$ ratio.
(iii) For the perception study, a separate set of menu models is trained on the same $70/30$
mixture; an RLBench-only \menu{RGB+D+N+S} model is included as unmatched context.
All groups are evaluated on RLBench tasks absent from training; ManiSkill3 serves only as a
training source.

\paragraph{Evaluation.}
Closed-loop control runs 20 trials per task with receding-horizon replanning under the
\texttt{IIII} plan, capped at $1.5\times$ the demonstration length; success is RLBench's
task-success signal, and methods run in our rollout harness are paired by scene seed. Future perception is scored
against the simulator state produced by each route's own decoded actions: LPIPS for RGB
\citep{zhang2018lpips}, absolute relative error for depth, mean cosine similarity for normals,
and mIoU for segmentation.

\subsection{Comparison with Robot Policies}
\label{sec:exp_external_policies}


\begin{table}[t]
\centering
\small
\setlength{\tabcolsep}{4.2pt}
\renewcommand{\arraystretch}{1.12}
\resizebox{0.85\textwidth}{!}{%
\begin{tabular}{@{}lcccccc@{}}
\toprule[1pt]
\textbf{Method}
& \shortstack{Close\\microwave}
& \shortstack{Toilet seat\\down}
& \shortstack{Reach\\target}
& \shortstack{Close\\drawer}
& \shortstack{Close\\laptop}
& \textbf{Avg.} \\
\midrule
PAD-Depth \citep{guo2024pad}
& 5 & 5 & 10 & 40 & 0 & 12 \\

$\pi_{0.5}$ \citep{physicalintelligence2025pi05}
& 30 & 5 & \textbf{35} & 35 & 25 & 26 \\

MolmoAct \citep{lee2025molmoact}
& 0 & 40 & 10 & 0 & 5 & 11 \\

UWM \citep{zhu2025uwm}
& 0 & 10 & 5 & 50 & 5 & 14 \\

LaWAM \citep{chen2026lawam}
& 15 & \textbf{70} & 10 & 25 & 5 & 25 \\

Evo-1 \citep{lin2026evo1}
& 10 & 10 & 0 & 10 & 5 & 7 \\

VLA-JEPA \citep{sun2026vlajepa}
& 0 & 20 & 5 & 30 & 15 & 14 \\

VLANeXt \citep{wu2026vlanext}
& 10 & 50 & 0 & 50 & 20 & 26 \\

\midrule
\textbf{XGenAct (ours)}
& \textbf{60} & 40 & 20 & \textbf{100} & \textbf{40} & \textbf{52} \\

\bottomrule[1pt]
\end{tabular}%
}
\caption{\small
\textbf{Closed-loop success rate (\%) on five unseen RLBench tasks}, 20 trials each.
All methods except MolmoAct are fine-tuned on our 16-task RLBench split; MolmoAct is evaluated
zero-shot with its output adapted to our controller, following \citet{zhen2026action}. Training and
evaluation details are in \appref{app:baselines}. Best result per column in bold.
}
\label{tab:policy-baselines}
\end{table}

We compare against eight policies fine-tuned on the same 16-task RLBench split (MolmoAct is
evaluated zero-shot, with its released checkpoint adapted to our controller as in Action Images;
\citealp{zhen2026action}): vision-language-action policies ($\pi_{0.5}$, Evo-1, VLANeXt), a VLA that
reasons over depth tokens (MolmoAct), and world action models that predict the future in pixel
space (PAD-Depth, UWM) or in a latent space (LaWAM, VLA-JEPA). All baselines except $\pi_{0.5}$
and LaWAM receive per-view camera parameters, since training cameras are randomized per episode.

As shown in \tabref{tab:policy-baselines}, \method{} reaches $52\%$ average success, twice the
strongest baselines ($26\%$ for $\pi_{0.5}$ and VLANeXt), and is best on three of the five tasks:
close drawer ($100\%$ vs.\ at most $50\%$), close microwave ($60\%$ vs.\ $30\%$) and close laptop
($40\%$ vs.\ $25\%$). The two other world action models that generate future pixels, PAD-Depth and
UWM, reach only $12\%$ and $14\%$. Baselines lead on toilet seat down (LaWAM, $70\%$) and reach
target ($\pi_{0.5}$, $35\%$), yet \method{} is the only method above $10\%$ on every task.

\begin{figure}[t]
  \centering
  \includegraphics[width=\linewidth]{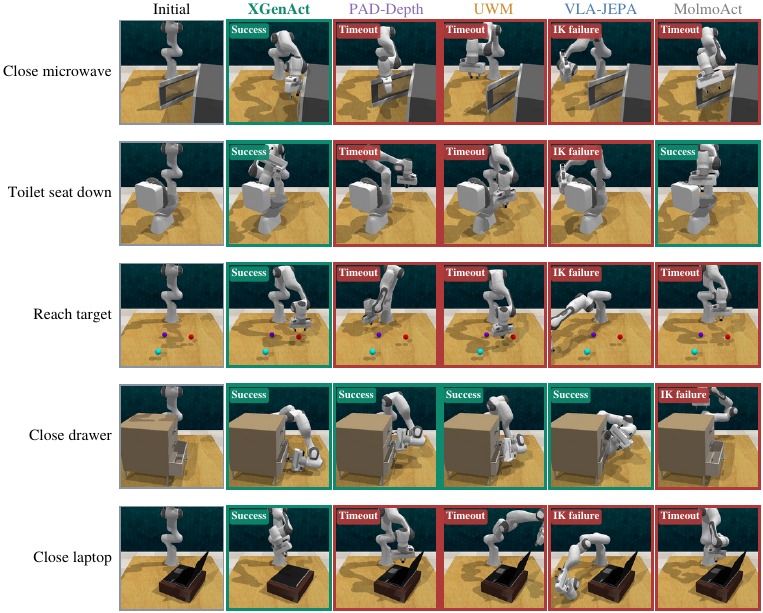}
  \caption{\small
  \textbf{Paired closed-loop rollouts against external policies.}
  Final frame of one episode per task and method; all methods start from the same scene
  (trials are paired by scene seed). The trial shown is the first one \method{} solves, and the
  baselines are shown on that trial whatever their outcome. Success rates over 20 trials are in
  \tabref{tab:policy-baselines}; per-task key frames and failure cases are in
  \appref{app:qualitative}.}
  \label{fig:closedloop_overview}
\end{figure}

The failure modes differ as well (Figure~\ref{fig:closedloop_overview}). Over 100 trials per
method, PAD-Depth and UWM mostly run out of time (88 and 86 timeouts), while VLA-JEPA and MolmoAct
often command poses the controller cannot reach and stop on inverse-kinematics (IK) failures (48 and
22 trials). \method{} never exhausts its stricter budget of 20 IK failures in total; every
failure is a timeout.

\subsection{Structured Perception Improves the Robot Control}
\label{sec:exp_ablation}

\begin{table}[t]
\centering
\small
\setlength{\tabcolsep}{3pt}
\renewcommand{\arraystretch}{1.00}
\begin{tabular}{@{}lccccc@{}}
\toprule[1pt]
\multirow{2}{*}{\textbf{Model}}
& \multicolumn{5}{c}{\textbf{RLBench}} \\
\cmidrule(lr){2-6}
& \shortstack{Close\\microwave}
& \shortstack{Toilet seat\\down}
& \shortstack{Close\\drawer}
& \shortstack{Meat on\\grill}
& \shortstack{RL\\avg.} \\
\midrule
\menu{RGB} & 5 & 20 & \textbf{95} & 10 & 32.50 \\
\menu{RGB+D} & 45 & 20 & \underline{85} & 60 & 52.50 \\
\menu{RGB+S} & \textbf{95} & 5 & 60 & \textbf{70} & 57.50 \\
\menu{RGB+N} & 75 & \textbf{40} & 70 & 45 & 57.50 \\
\menu{RGB+D+S} & 70 & 25 & 60 & 60 & 53.75 \\
\menu{RGB+N+S} & 80 & \underline{35} & 65 & \underline{65} & \textbf{61.25} \\
\menu{RGB+D+N} & 75 & \underline{35} & 60 & 55 & 56.25 \\
\menu{RGB+D+N+S} & \underline{85} & \underline{35} & 55 & 60 & \underline{58.75} \\
\bottomrule[1pt]
\end{tabular}
\caption{\small\textbf{Closed-loop success (\%) across training-time modality menus.}
Bold and underlined values are the best and second-best results per column.}
\label{tab:unseen_success}
\end{table}

\tabref{tab:unseen_success} compares the eight models of group (ii), which differ only in their
training menu. Every menu with a perception stream
improves on \menu{RGB}, raising average success from $32.5\%$ to $52.5$--$61.25\%$; the best,
\menu{RGB+N+S}, gains $28.75$ points. The most useful stream depends on the task.
Segmentation helps most where the manipulated part must be separated from its surroundings:
\menu{RGB+S} is best on close microwave ($95\%$) and meat on grill ($70\%$). Normals help
rotating a thin planar part: \menu{RGB+N} is best on toilet seat down ($40\%$), and the three
runners-up all include normals. Close drawer is the exception, where \menu{RGB} already reaches
$95\%$.
The results show that more perception tasks is not monotonically better. Adding depth to \menu{RGB+N+S} lowers the average
from $61.25\%$ to $58.75\%$, although depth alone is clearly useful ($52.5\%$ for \menu{RGB+D}).
Under a fixed model capacity and training budget, the streams therefore overlap or compete, and
the best menu is the one whose geometry or semantics resolve a task's main ambiguity. These trends
are correlational; paired executions of \menu{RGB} and \method{} on the same scenes are shown in
Figure~\ref{fig:paired_rollouts}.

\subsection{Future Perception under Simulator Replay}
\label{sec:exp_main}

Every \method{} model can produce a future perception map in two ways: generate it directly with
the corresponding template (e.g.\ \tmpl{depth+action}), or generate future RGB with
\tmpl{video+action} and pass it to a frozen expert (Depth Anything V2 for depth, Lotus-G for
normals, CLIPSeg for segmentation). We compare the two routes on four held-out tasks using the
perception-study models (group~iii). Since the robot executes each route's own decoded actions,
both are scored against the simulator state those actions reach, not against a demonstration the
rollout may have left (Figure~\ref{fig:eval_pipeline}).

\setlength{\textfloatsep}{10pt}
\begin{figure}[t]
  \centering
  \includegraphics[width=\textwidth]{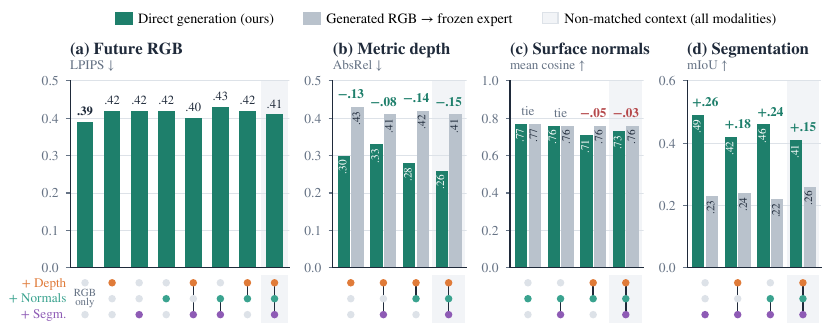}
  \caption{\small
  \textbf{Direct generation versus generated RGB followed by a frozen expert.}
  Each column is one training menu; the dots below mark the perception streams it adds to RGB.
  Teal bars: the model generates the perception space directly. Gray bars: the same model
  generates future RGB, which Depth Anything V2, Lotus-G or CLIPSeg converts. Numbers above a
  pair give direct minus expert (green: direct is better). Scores average four held-out RLBench
  tasks, measured against the simulator state reached by each
  route's own actions; a task is left out of a panel if its executed poses deviate from the
  commanded ones by 10\,mm or more (\appref{app:protocol}; per-task values in \tabref{tab:main}).
  }
  \label{fig:percep_v3}
\end{figure}

Figure~\ref{fig:percep_v3} shows that generating geometry and semantics directly is clearly
better than reading them off generated RGB by external experts: depth AbsRel drops from $0.41$--$0.43$ to
$0.28$--$0.33$, and segmentation mIoU roughly doubles, from $0.22$--$0.24$ to $0.42$--$0.49$. For
surface normals the two routes are on par ($0.76$--$0.77$ cosine), except with \menu{RGB+D+N},
where the direct route trails ($0.71$ vs.\ $0.76$). Adding perception streams costs little in
future RGB: LPIPS rises only from $0.39$ for \menu{RGB} to $0.40$--$0.43$. The RLBench-only
\menu{RGB+D+N+S} model, shown as unmatched context, follows the same pattern ($0.26$ AbsRel,
$0.41$ mIoU, $0.73$ cosine).

\begin{figure}[t]
  \centering
  \includegraphics[width=\linewidth]{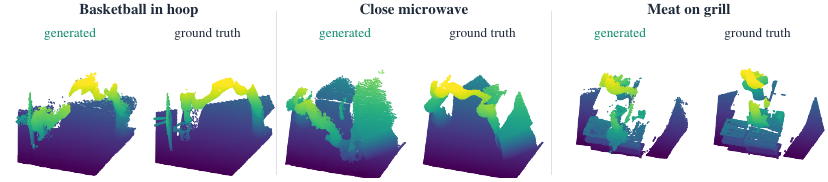}
  \caption{\small
  \textbf{Generated metric depth lifted to 3-D on held-out tasks.} The last generated depth frame
  (view~1, open loop) and the demonstration's depth at the same frame are back-projected with the
  camera intrinsics and rendered from the same novel viewpoint and axis limits; colour encodes
  height. Pixels on depth discontinuities are dropped in both clouds. Further qualitative results
  are in \appref{app:qualitative}.}
  \label{fig:pointclouds}
\end{figure}

Because the depth stream is metric, a generated frame can be used directly as geometry.
Figure~\ref{fig:pointclouds} back-projects the last generated depth frame on three held-out tasks: the
table, fixtures and object layout coincide with the ground-truth cloud, and the visible differences
lie on the arm, whose predicted pose runs ahead of or behind the demonstration.

The imagined future also matches what the robot then does. Appendix~\ref{app:qual-imagination}
pairs generated RGB frames with simulator frames reached by executing their decoded actions;
the plan is carried out as imagined rather than replaced by a different motion at execution time.


\section{Conclusion}
\label{sec:conclusion}

We presented \method{}, which expresses robot actions, future RGB video, metric depth, surface
normals, and functional role segmentation through one video-generation interface. Deterministic
codecs let these tasks share a frozen video VAE, diffusion transformer, output head, and objective,
without modality-specific learned heads or auxiliary losses. On held-out RLBench tasks, structured
perception improves closed-loop success over RGB-only training, while direct generation predicts
future depth and segmentation more accurately than the evaluated RGB-to-expert pipelines. The
best training menu remains task-dependent, motivating broader evaluations and adaptive task
balancing.

\newpage
\section*{AI Use Statement}

Generative AI tools were used for literature retrieval and discovery, research ideation, and
experimental support. During early brainstorming, we began with the broad idea of transferring
Vision Banana to robotic manipulation. Generative AI tools were used to search for and summarize
related literature on world action models, which helped us identify relevant prior work, including
Action Images. During research execution, these tools assisted with understanding, implementing,
and debugging baseline methods. They also served as a discussion aid for conditioning strategies,
ablation studies, and experimental design, and helped draft preliminary experiment plans. During
paper writing, generative AI was used to draft parts of the related work, experimental setup,
tables, and appendix; to polish the method and introduction; and to help format the manuscript.

All AI-assisted research activities and generated content were
reviewed and verified by the authors. The authors take responsibility
for the final methodology, experiments, results, and manuscript.

\section*{Reproducibility Statement}

We provide the information needed to independently reproduce our method and
experiments throughout the paper and appendix. Section~\ref{sec:method} and
Appendix~\ref{app:prelim} specify the model, objective, and action
representation; Appendices~\ref{app:data}--\ref{app:camera} document data
construction and disjoint splits, deterministic modality codecs and admission
checks, the complete training mixture, optimization hyperparameters, and camera
conventions; and Appendices~\ref{app:protocol} and~\ref{app:baselines} define
the evaluation and external-baseline protocols. We will publicly release the
complete code for data generation and preprocessing, training, inference,
closed-loop rollout, and evaluation, together with pinned environment
specifications, launch commands, configuration files, task and split
definitions, and the random seeds used in the reported experiments. We will
also release the trained checkpoints for every reported \method{} variant.
For simulator assets and third-party checkpoints governed by their own
licenses, we will provide versioned download and reconstruction instructions
rather than redistribute them. The release will map every table and figure to
its configuration and checkpoint, with scripts that reproduce the reported
metrics from raw simulator outputs.



\newpage
\bibliography{references}
\bibliographystyle{xgenact_preprint}

\newpage

\appendix
\section{Architecture and action representation}
\label{app:prelim}

\subsection{Latent video-diffusion world action model}
\label{app:backbone}

We build on Wan2.2-TI2V-5B \citep{wan2025}, a text-conditioned latent video
diffusion transformer whose causal 3D VAE encoder and decoder, $\gE$ and $\gD$,
stay frozen throughout.
A clip of $T$ frames at resolution $H\times W$ is encoded as
\begin{equation}
    \gE:
    \R^{3\times T\times H\times W}
    \longrightarrow
    \R^{C_l\times T_l\times H_l\times W_l},
    \qquad
    T_l = 1+\frac{T-1}{4},
    \quad
    H_l=\frac{H}{16},
    \quad
    W_l=\frac{W}{16}.
    \label{eq:vae}
\end{equation}
For our $41$-frame, $512^2$ clips this is an $11\times32\times32$ grid with
$C_l=48$ channels.
Only the diffusion transformer $f_{\vtheta}$ is fine-tuned.
Each sample carries $V=2$ camera views, conditioned as in
\citet{zhen2026action} through Pl\"ucker ray embeddings
\citep{sitzmann2021light,bai2025recammaster}, collected in a tensor $\tC$
(\appref{app:camera}).

\subsection{Actions as video}
\label{app:actionimage}

Following Action Images \citep{zhen2026action}, actions live in the same
RGB-video space as observations.
For an end-effector state $(\vp_n,\mR_n,g_n)$ at frame $n$ --- position,
orientation, gripper openness --- view $v$ projects three image points,
\begin{equation}
    \vq^{(1)}_n = \pi_v(\vp_n),
    \qquad
    \vq^{(c)}_n
    =
    \pi_v\!\left(
        \vp_n+\ell\,\mR_n\ve_c
    \right),
    \quad
    c\in\{2,3\},
    \label{eq:actionpoints}
\end{equation}
with fixed tool length \(\ell=0.1\) m and tool-frame axes
\(\ve_2=(1,0,0)^\top\) and \(\ve_3=(0,0,-1)^\top\).
They are rendered into three RGB channels as Gaussian blobs,
\begin{equation}
    \etA^{(v)}_{c,n,\vx}
    =
    \exp\!\left(
        -\frac{
            \|\vx-\vq^{(c)}_n\|_2^2
        }{2\sigma^2}
    \right),
    \qquad
    \sigma=0.05\min(H,W).
    \label{eq:actionimage}
\end{equation}
To the frozen VAE and the transformer the action stream is therefore an
ordinary RGB video.
Gripper encoding and the trajectory decoder follow \citet{zhen2026action}
(\appref{app:codecs}).

\section{Training data and evaluation splits}
\label{app:data}

\begin{table}[ht]
\centering
\small
\setlength{\tabcolsep}{5pt}
\renewcommand{\arraystretch}{1.15}
\begin{tabularx}{\textwidth}{@{}lX@{}}
\toprule
\textbf{Source} & \textbf{Training tasks} \\
\midrule
RLBench &
\texttt{open\_drawer}, \texttt{slide\_block\_to\_target},
\texttt{sweep\_to\_dustpan}, \texttt{meat\_off\_grill}, \texttt{turn\_tap},
\texttt{put\_item\_in\_drawer}, \texttt{close\_jar}, \texttt{reach\_and\_drag},
\texttt{stack\_blocks}, \texttt{light\_bulb\_in}, \texttt{put\_money\_in\_safe},
\texttt{stack\_wine}, \texttt{put\_groceries\_in\_cupboard},
\texttt{place\_shape\_in\_shape\_sorter}, \texttt{push\_buttons},
\texttt{insert\_onto\_square\_peg} \\
\addlinespace
ManiSkill3 &
\texttt{peg\_insertion\_side}, \texttt{pick\_cube}, \texttt{plug\_charger},
\texttt{pull\_cube\_tool}, \texttt{push\_cube}, \texttt{stack\_cube},
\texttt{stack\_pyramid} \\
\bottomrule
\end{tabularx}
\caption{Training tasks of the two sources. The external comparison trains on the RLBench tasks
only; the menu ablation and the perception study train on both sources at a $70/30$ ratio.
Names follow the identifiers used by the data loader.}
\label{tab:training_tasks}
\end{table}

\paragraph{RLBench source.}
The RLBench source contains 16 tasks, four camera views (front, overhead, left shoulder, and
right shoulder), and $512^2$ observations recorded at 20\,Hz,
with per-frame metric depth, instance masks, and camera intrinsics and extrinsics written
alongside every episode. Variation~0 is the only split used for training and holds $3{,}960$
episodes. We targeted 250 seeds per task; \texttt{light\_bulb\_in} and
\texttt{place\_shape\_in\_shape\_sorter} yielded 249 each, and
\texttt{sweep\_to\_dustpan} yielded 212. Every episode is regenerated from its seed
under Colosseum-style randomization \citep{pumacay2024colosseum} of camera pose, table
color, table texture, background texture, and light color. Object color, object texture,
distractors, and object physics are deliberately \emph{not} randomized: color is part of the
instruction semantics (``close the \emph{red} jar''). The randomization seed is derived from
\texttt{crc32(task|variation|seed)} rather than from Python's hash, so a tree is
reproducible pixel-for-pixel across processes.

\paragraph{ManiSkill3 source.}
The ManiSkill3 source contains seven tasks and $1{,}750$ training episodes. We produce it by
replaying the official ManiSkill demonstrations \citep{taomaniskill3} frame by frame through
\texttt{set\_state\_dict} and re-rendering them, rather than by collecting new trajectories.
Each task contributes 250 episodes, and each episode is rendered from
four cameras sampled on a sphere around the workspace center, at $512^2$ and 20\,Hz, with one
rendered frame per control step. Camera parameters are converted from the OpenCV convention
to the RLBench convention of \appref{app:camera}, and each episode is admitted only if
unprojecting and reprojecting its own rendered depth closes to under $0.51$\,px, if no mask
pixel falls outside the segmentation vocabulary, and if the fraction of pixels outside the
depth codec range stays under $3\%$; failures resample the cameras and, if needed, retry the
episode. The byte-level layout is identical to the RLBench tree, so both sources are read by
the same loader.

\paragraph{Mixture.}
For the menu ablation and the perception study, training draws a source with probability
$0.7$ for RLBench and $0.3$ for ManiSkill3, then an episode uniformly within that source. Over the $24{,}000$ samples of a 6k-step run this
gives $4.24$ exposures per RLBench episode and $4.11$ per ManiSkill3 episode, and
$4.4\%$ versus $4.3\%$ of samples per task --- so the two sources are balanced both per
episode and per task rather than only in aggregate. Windows are 41 frames; the RLBench
source uses stride 3, which covers $6.0$\,s at an effective $6.7$\,Hz, and the ManiSkill3
source uses stride 1, $2.0$\,s at 20\,Hz, because its episodes are only 49--103 frames long
and a larger stride would degenerate into frozen tail frames.

\paragraph{Evaluation suites.}
No evaluation episode is drawn from either training tree. The perception study in
\tabref{tab:main} uses four RLBench tasks absent from training:
\texttt{close\_microwave}, \texttt{toilet\_seat\_down}, \texttt{close\_box}, and
\texttt{meat\_on\_grill}. Each task is evaluated over five simulator rollouts at training
step~4{,}000. The internal closed-loop ablation in \tabref{tab:unseen_success} uses
\texttt{close\_microwave}, \texttt{toilet\_seat\_down}, \texttt{close\_drawer}, and
\texttt{meat\_on\_grill}, with 20 trials per task. The external-policy comparison in
\tabref{tab:policy-baselines} evaluates \texttt{close\_microwave},
\texttt{toilet\_seat\_down}, \texttt{reach\_target}, \texttt{close\_drawer}, and
\texttt{close\_laptop\_lid}, also with 20 trials per task. The three protocols use the three
separately trained groups of Section~\ref{sec:exp_setup} and should not be pooled.

\paragraph{Fresh closed-loop scenes.}
The rollout harness generates a fresh demonstration per trial from a seed derived from
\texttt{(task, variation, trial)}, disjoint from the seeds used to render either training
tree, so a closed-loop trial is a new scene rather than a replay of a stored episode.

\section{Modality codecs and admission-gate fidelity}
\label{app:codecs}

\paragraph{Depth decoding.}
For a valid metric depth \(d\in(0.05,10)\) meters, we first apply the monotone transform
\begin{equation}
    y(d)
    =
    1-
    \left(
        1-\frac{d}{\lambda c}
    \right)^{\lambda+1},
    \qquad
    \lambda=-3,
    \quad
    c=\frac{10}{3}.
    \label{eq:depth-transform}
\end{equation}
which is the power transform of \citet{barron2025power}. We then trace a piecewise-linear
Hamiltonian path through the eight vertices of the RGB cube, following \citet{gabeur2026image}:
\begin{equation}
    \gamma_{\mathrm{depth}}(d)
    =
    (1-\alpha)\mathbf{v}_k+\alpha\mathbf{v}_{k+1},
    \qquad
    k=\lfloor 7y(d)\rfloor,
    \quad
    \alpha=7y(d)-k,
    \label{eq:depthcodec}
\end{equation}
where consecutive vertices differ in one RGB coordinate.
At decoding time, each generated RGB pixel is projected onto its nearest
path edge.
The position along the path recovers \(y\), after which the Barron
transform is inverted to obtain metric depth.
Pixels farther than a fixed tolerance from every path edge are marked
\emph{invalid} instead of being assigned an arbitrary depth value.
This provides an explicit detector for generations that leave the valid
measurement manifold.

\paragraph{Surface normals.}
Normals are computed from metric depth using camera-aware finite differences:
\[
    \widetilde{\vn}
    =
    \left(
        -\frac{\partial d}{\partial u}\frac{|f_x|}{d},
        -\frac{\partial d}{\partial v}\frac{|f_y|}{d},
        1
    \right),
    \qquad
    \vn=\frac{\widetilde{\vn}}{\|\widetilde{\vn}\|_2}.
\]
Each unit normal is then encoded as
\begin{equation}
    \gamma_{\mathrm{normal}}(\vn)=\frac{\vn+\mathbf{1}}{2}.
    \label{eq:normal}
\end{equation}
Decoding first maps RGB values back to $[-1,1]^3$ and then renormalizes the
result to unit length.

\paragraph{Segmentation.}
Segmentation uses a fixed lookup table from functional-role labels to RGB values, shared
across tasks, episodes, frames, and camera views. The eight training roles are
\texttt{background}, \texttt{target}, \texttt{goal}, \texttt{robot\_arm},
\texttt{gripper}, \texttt{distractor}, \texttt{tool}, and \texttt{fixture}. A ninth
\texttt{unknown} color is reserved as an annotation diagnostic and is rejected during data
validation rather than used as a training class.
The labels are functional roles a task assigns --- what is being manipulated,
what it is being manipulated towards, what merely populates the scene --- so
one color keeps one meaning across every task in the tree.
Unlike a referring-mask representation, the prompt does not specify a per-episode color map;
each color has a stable functional interpretation throughout training.

\paragraph{Action images.}
For the action representation in \eqref{eq:actionimage}, the gripper center and two
orientation-axis endpoints occupy the three RGB channels.
Following \citet{zhen2026action}, the low-response background of the blue channel additionally
stores a binary gripper-openness pedestal.
Decoding extracts the image-space peaks in each camera and fuses the
corresponding rays to recover a 3D end-effector trajectory.

\begin{table}[!ht]
\centering
\caption{
Admission-gate results for each output stream.
All streams are evaluated after a codec round-trip and, where applicable,
a round-trip through the frozen VAE before training.
Metrics are computed against the decoded ground-truth stream.
The RGB control uses the same VAE and clips and provides an approximate
upper bound on reconstruction quality.
}
\label{tab:gate}

\begin{threeparttable}
\small
\setlength{\tabcolsep}{4pt}
\renewcommand{\arraystretch}{1.12}

\begin{tabularx}{\linewidth}{
    @{}
    l
    >{\raggedright\arraybackslash}X
    >{\raggedright\arraybackslash}p{0.19\linewidth}
    >{\raggedright\arraybackslash}p{0.27\linewidth}
    @{}
}
\toprule
\textbf{Stream}
& \textbf{Encoding}
& \textbf{Codec round-trip}
& \textbf{VAE round-trip} \\
\midrule

Depth
& Barron $\rightarrow$ RGB-cube path, \eqref{eq:depthcodec}
& AbsRel $0.066\%$
& AbsRel \textbf{0.362\%}; $0.02\%$ off-path \\

Normal
& $\tfrac{1}{2}(\vn + 1)$, \eqref{eq:normal}
& mean cosine $0.999998$
& mean cosine \textbf{0.9906} ($7.9^\circ$) \\

Segmentation
& Role LUT with a 9-color palette
& IoU $>0.98$ (spec.)
& \texttt{target} IoU median \textbf{0.965} \\

Action
& Three Gaussian channels, \eqref{eq:actionimage}
& $\approx 4$\,mm 3D
& Not measured\tnote{$\dagger$} \\

\midrule

\multicolumn{2}{@{}l}{\textit{RGB control (same VAE, same clips)}}
& N/A
& 29.6--31.5\,dB PSNR \\

\bottomrule
\end{tabularx}

\begin{tablenotes}[flushleft]
\scriptsize
\item[$\dagger$]
The action stream is rendered online rather than loaded from disk, so an equivalent
VAE admission test requires a separate action-decoding harness and is not reported here.
\end{tablenotes}

\end{threeparttable}
\end{table}

\section{Training-mixture details}
\label{app:mixture}

The training distribution factorizes as
\[
    p(\Pi,\rho)
    =
    p(\Pi)\,p(\rho\mid\Pi),
\]
where $\Pi$ determines the modality template and $\rho$ the conditioning plan.

\paragraph{Template distribution.}
\method{} draws
\tmpl{video+action} with probability $0.4$ and each of
\tmpl{depth+action}, \tmpl{segmentation+action}, and \tmpl{normal+action} with
probability $0.2$.
The RGB-only control draws \tmpl{video+action} with probability \(1.0\).
Every template in either mixture carries an action stream, and every template
has the same four-segment shape and the same cost, so the two runs differ in
the sampled observation space while retaining the same action-supervision budget.

\paragraph{Conditioning plans.}
An action stream is dropped with probability \(0.1\) before the plan is drawn.
Such a draw becomes a single-modality sample conditioned on first frames only:
generate the whole depth, segmentation, normal or RGB video from its own first
latent frame. The remaining \(90\%\) of draws use the conditioning distribution of
\tabref{tab:action-mixture}. The \texttt{1I1I} plan is applied only to RLBench samples; a
ManiSkill3 sample falls through to the other two plans, which gives it \(93.75\%\)
\texttt{IIII} and \(6.25\%\) \texttt{FIFI}.

\begin{table}[!ht]
\centering
\caption{
The shared conditioning distribution used by every trained model, including \menu{RGB},
conditional on the action stream being retained (RLBench samples).
}
\label{tab:action-mixture}
\small
\begin{tabular}{lccc}
\toprule
& \texttt{IIII} & \texttt{1I1I} & \texttt{FIFI}
\\
\midrule
Conditional & $75\%$ & $20\%$ & $5\%$  \\
\bottomrule
\end{tabular}
\end{table}

The conditioning-plan distribution, optimization schedule, and action dropout are therefore
matched across all runs; their training distributions differ only in the template menu.

\section{Flow-matching and optimization details}
\label{app:optimization}

We retain the upstream flow-matching time sampler and weighting rule.
A base time \(\hat t\) is drawn uniformly from the discrete \(1000\)-point
schedule over $(0,1]$ and shifted according to
\begin{equation}
    t
    =
    \frac{s\hat t}{
        1+(s-1)\hat t
    },
    \qquad
    s=5.
    \label{eq:timeshift}
\end{equation}
The timestep weight \(w(t)\) in \eqref{eq:loss} is obtained by subtracting the minimum
of \(\exp[-2(t-\tfrac{1}{2})^2]\) over the discrete schedule and normalizing the resulting
weights to have unit mean.

Unless otherwise stated, all runs warm-start from the released
Action Images checkpoint at step $125{,}750$.
We fine-tune all parameters of the diffusion transformer (the 5B Wan2.2 backbone plus the
per-block camera modules, 6.4B parameters in total) in bfloat16 mixed precision with
ZeRO-2 and CPU-offloaded Adam, gradient checkpointing, batch size $1$ per
device on four GPUs and no accumulation --- an effective batch of 4 --- at
learning rate $5\times10^{-7}$ with $1000$ warmup steps, a constant schedule
thereafter, gradient clipping at $1.0$, and seed 42.
The matched mixed-source runs use 6{,}000 optimization steps and save a checkpoint every
1{,}000 steps; the reported perception comparison reads all matched checkpoints at
step~4{,}000.
The VAE remains frozen throughout.

\section{Camera conditioning and coordinate conventions}
\label{app:camera}

Both data sources store, per view and per frame, a $4\times4$ camera-to-world
extrinsic matrix and a $3\times3$ intrinsic matrix in the RLBench convention,
whose camera frame has $x$ to the left, $y$ up and $z$ forward, so that $f_x$
and $f_y$ carry a negative sign.
ManiSkill3 reports OpenCV parameters instead, which we convert once at
generation time,
\[
    C_{\mathrm{rlb}}
    =
    E_{\mathrm{cv}}^{-1}\,
    \mathrm{diag}(-1,-1,1,1),
    \qquad
    K_{\mathrm{rlb}}
    =
    K_{\mathrm{cv}}\,
    \mathrm{diag}(-1,-1,1),
\]
so that a single projection routine serves both sources and the action codec of
\eqref{eq:actionpoints} needs no per-source branch.
Camera conditioning itself follows \citet{zhen2026action}: the per-view rays, relative to the
first frame of view~1, are embedded as Pl\"ucker coordinates
\citep{sitzmann2021light,bai2025recammaster} and encoded by the frozen VAE into the tensor $\tC$.
Each transformer block has its own camera encoder, whose output is added to the token features
before self-attention; every segment reuses the camera of its view. The absolute extrinsics and
intrinsics are used only to project and decode action images.

\section{Evaluation protocol}
\label{app:protocol}

\paragraph{Sampling configuration.}
The matched \method{} checkpoints use 50 denoising steps, classifier-free guidance \(7.5\),
explicit modality tags, temporal stride 3, and seed 42. Their action images are decoded with
the same sphere-constrained two-view solver.

\begin{figure}[t]
  \centering
  \includegraphics[width=\textwidth]{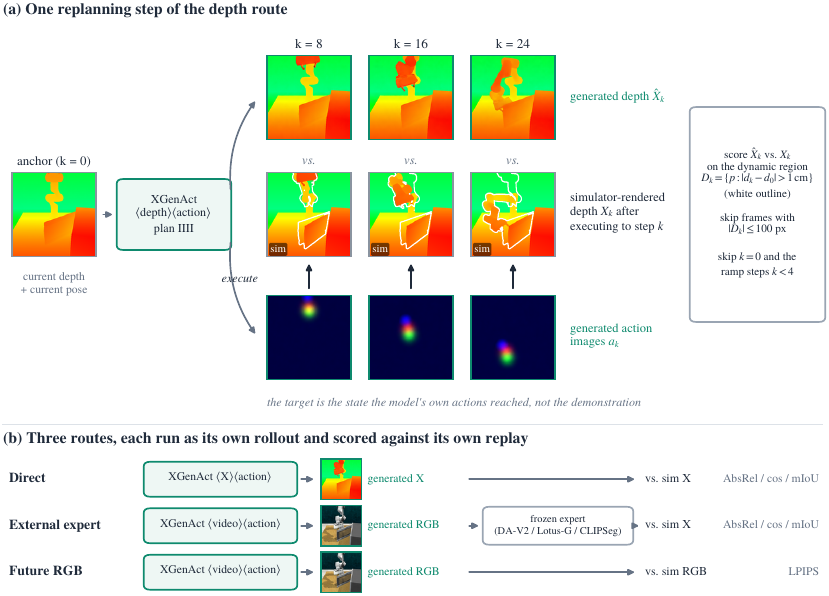}
  \caption{\small
  \textbf{Perception evaluation by simulator replay.}
  \textbf{(a)} One replanning step of the depth route (close microwave, view~1). From the current
  depth and pose, \method{} generates a chunk of depth frames $\hat X_k$ and action images $a_k$.
  The decoded poses are executed, and after each step $k$ the simulator renders the depth $X_k$ of
  the state that the model's own actions reached (middle row, raw simulator output in the same
  colour code); $\hat X_k$ is scored against $X_k$ rather than against a demonstration. Scores use
  only the dynamic region $D_k$, outlined in white: pixels whose simulator depth changed by more
  than 1\,cm since the chunk began, which includes both where the arm is and where it was. The
  anchor $k=0$ and the ramp steps $k<4$ are never scored.
  \textbf{(b)} The three routes compared in \tabref{tab:main} and Figure~\ref{fig:percep_v3}.
  Each is a separate rollout scored against its own replay: direct generation of $X$, the same
  checkpoint's generated RGB passed through a frozen expert, and generated RGB itself.
  }
  \label{fig:eval_pipeline}
\end{figure}

\paragraph{Future perception and the dynamic region.}
A perception space of this model is only ever prompted alongside an action, so under
\texttt{IIII} the anchor is frame~0 of that same modality and the scored frames are the ones
that follow it.
The baseline that carries the comparison is the \emph{external-expert route}: the checkpoint's own
\tmpl{video+action} stream is sampled under the identical configuration, and a feed-forward
specialist is then run on the frames it generated.
The reported depth route uses the raw metric output of Depth Anything V2 Metric Indoor-Large
with no fitted alignment parameters.
Normals are scale-free and receive no fit; for segmentation the external expert is given the label
vocabulary of the episode.
\emph{Copy-anchor} repeats the anchor at every future time and is retained as a protocol
diagnostic, but is omitted from \tabref{tab:main} for readability.
The first four decoded poses in each action chunk are replaced during execution by a ramp from
the current pose to the fifth pose. Since these ramp poses are not predictions made by the
model, scoring excludes them as well as the anchor at $k=0$.
For every modality, the \emph{dynamic region} is defined from simulator depth as
\[
    \mathcal{D}_t=\{p:\lvert d_t(p)-d_0(p)\rvert>0.01\ \mathrm{m}\},
\]
where \(d_0\) is the simulator depth after the first executed step of the current chunk.
The same region is used for RGB, depth, normals, and segmentation, and a frame is assigned a
dynamic-region score only when \(\lvert\mathcal{D}_t\rvert>100\) pixels. Because this mask
depends only on simulator ground truth, no prediction can influence which pixels are scored.
Restricting evaluation to \(\mathcal{D}_t\) prevents static background from dominating the
metric.

\paragraph{Main-table reporting details.}
Direct rows in \tabref{tab:main} appear only when the scored modality is present in the training
mixture. For every checkpoint, the external-expert route uses its RGB output and then applies
Depth Anything V2 Metric Indoor-Large, Lotus-G, or CLIPSeg. Future RGB has no
external-expert route because
it is already the intermediate representation consumed by the other cascades. It is scored
against lossless simulator RGB dumped after every executed step, never against a demonstration
future or an MP4-decoded frame. Table~\ref{tab:main} reports RLBench evaluation only; ManiSkill3 contributes training data
through the fixed \(30\%\) source share. A metric block excludes a task from every route's average whenever any shown cell in that
block has a median commanded--achieved end-effector gap of at least 10\,mm; the affected cells
remain visible and are marked
$^{\dagger}$. Every task cell uses five trials.

\paragraph{Segmentation metrics.}
\textbf{mIoU} is macro-averaged over the named labels present in the scored region. Because the
live simulator annotations do not identify which distractor the instruction promotes to
\texttt{target}, \texttt{target} is merged into \texttt{distractor} in both prediction and
ground truth before scoring.

\paragraph{Closed loop.}
RLBench uses its rollout harness with receding-horizon replanning capped at
$1.5\times$ the demonstration length and is evaluated with the corresponding control checkpoint.

\section{External policy baselines}
\label{app:baselines}

This appendix describes how the external policies in \tabref{tab:policy-baselines} are
trained and evaluated. Every baseline runs from its official code base. Except for MolmoAct,
which is evaluated zero-shot and adapted to our interface as in Action Images
\citep{zhen2026action}, each is fine-tuned on the 16 RLBench training tasks of
\appref{app:data}, variation~0; none of the five evaluation tasks appears in this split.
\tabref{tab:baseline-config} summarizes the settings.

\begin{table}[ht]
\centering
\small
\setlength{\tabcolsep}{4pt}
\renewcommand{\arraystretch}{1.12}
\begin{tabular}{@{}llcccl@{}}
\toprule
\textbf{Method} & \textbf{Training} & \textbf{Cam.} & \textbf{Chunk} & \textbf{Exec.}
& \textbf{Action target} \\
\midrule
$\pi_{0.5}$ & 20{,}500 steps, batch 32 & $\times$ & 50 & 10 & $\Delta$SE(3) to current pose \\
LaWAM       & 20{,}500 steps, batch 32 & $\times$ &  8 &  8 & absolute pose \\
Evo-1       & 20{,}500 steps, batch 32 & \checkmark & 50 & 25 & per-step delta, accumulated \\
VLANeXt     & 20{,}500 steps, batch 32 & \checkmark &  8 &  8 & absolute pose \\
\midrule
MolmoAct    & zero-shot                & \checkmark & 16 & 16 & triangulated 2D trace \\
PAD-Depth   & 2{,}560 steps, batch 256 & \checkmark & 3 poses & 16 & absolute pose \\
UWM         & 4{,}600 steps, batch 36  & \checkmark & 16 &  8 & $\Delta$ to current pose \\
VLA-JEPA    & 2{,}000 steps, batch 32 & \checkmark &  7 &  4 & absolute pose \\
\midrule
\method{}   & 4{,}000 steps, batch 4   & \checkmark & 41 & 41 & action image \\
\bottomrule
\end{tabular}
\caption{\small
Training and execution settings of the methods in \tabref{tab:policy-baselines}.
\emph{Cam.}: camera parameters are an input. \emph{Chunk}: actions predicted per query;
\emph{Exec.}: actions executed before replanning.}
\label{tab:baseline-config}
\end{table}

\paragraph{Camera parameters.}
The training tree randomizes camera poses per episode (\appref{app:data}), whereas rollouts
use RLBench's fixed default cameras. \method{} receives the camera through Pl\"ucker
embeddings and its action projection (\appref{app:camera}). A baseline marked in
\tabref{tab:baseline-config} receives, per view, the $3\times4$ extrinsic matrix and the
intrinsics $(f_x,f_y,c_x,c_y)$ rescaled to its input resolution, 16 values per view.
The $\pi_{0.5}$ and LaWAM rows are camera-free variants.

\paragraph{$\pi_{0.5}$, LaWAM, Evo-1, and VLANeXt.}
These four policies share one training budget: $20{,}500$ steps at an effective batch size of
32 on the 16-task tree ($3{,}960$ episodes, $690{,}893$ frames at 20\,fps), with $512^2$
renders resized to $224^2$. Each executes its action chunks as its official inference code
does. $\pi_{0.5}$ predicts 50 SE(3) deltas relative to the current pose and executes 10.
LaWAM and VLANeXt predict 8 absolute poses and execute the whole chunk. VLANeXt conditions on
an 8-frame proprioceptive history, which its server rebuilds from the chunks it returned,
because the client state is the last commanded pose. We load its weights with zero missing
or unexpected keys; it keeps no EMA copy. Evo-1 executes the first 25 of 50 actions, as in
both of its official clients. Its targets are per-step deltas, since each state in the data
equals the previous commanded action, so a chunk is composed by accumulating the deltas.
Its rotation deltas are computed by subtracting rotation vectors, which wrap at $\pm\pi$;
$2.06\%$ of training frames therefore carry a rotation delta near $2\pi$. The reported Evo-1
result consequently reflects this data-conversion defect and should not be interpreted as an
estimate of an otherwise correctly trained Evo-1 policy's capability.

\paragraph{MolmoAct.}
Following \citet{zhen2026action}, we evaluate the released MolmoAct-7B-D checkpoint zero-shot with
the reasoning prompt from its model card and adapt its output to our controller as follows. MolmoAct draws a 2D end-effector trace on its primary input image, so each replan
queries it twice, once with each view as the primary image. The two traces are resampled by
arc length to 16 corresponding points, triangulated with the true camera matrices, and
anchored to the current end-effector position. Orientation is held at its current value; the
gripper command is the mean of the two predicted gripper values, thresholded at 0.5. All 16
waypoints are executed before replanning.

\paragraph{PAD-Depth.}
PAD-Depth starts from the released Bridge-pretrained DiT-XL/2 (681M parameters). It uses the
frozen SD VAE and CLIP text encoder and is fine-tuned on the \texttt{front} view at $256^2$
together with metric depth and the front camera parameters. It predicts $k=3$ future frames
and poses spaced 16 control steps apart, at global batch size 256 and learning rate
$10^{-4}$. We evaluate the step-$2{,}560$ checkpoint, which has seen $655{,}360$ samples, the
same exposure as the $20{,}500\times32$ budget above. We fixed three issues in the released
training code. The depth loader referenced an undefined attribute. The action loss was
disabled for the first $50{,}000$ steps; we enable it from step~0 with weight 1. The first
future target was the current frame; we shift targets forward by one stride. Sampling uses
75 DDIM steps, as in the paper. Following the paper, only the first predicted pose is
executed; it is linearly interpolated (orientation by slerp) into 16 waypoints to match the
step budget of other policies.

\paragraph{UWM.}
UWM starts from the released DROID-pretrained checkpoint and is fine-tuned for $4{,}600$ steps
at batch size 36 and learning rate $10^{-4}$ with delta actions. It observes the
\texttt{front} and \texttt{overhead} views at $128^2$ over two frames, together with a
42-dimensional state: a 10-dimensional pose (position, 6D rotation, gripper) and 16 camera
values per view. It predicts 16 actions and executes 8.

\paragraph{VLA-JEPA.}
VLA-JEPA starts from the released DROID-pretrained checkpoint (\texttt{VLA-JEPA-Pretrain}) and is
fine-tuned with its LeRobot training code. Our actions are 10-dimensional poses and our state is
the 42-dimensional vector used for UWM, whereas the pretrained model uses 7- or 8-dimensional
ones, so the action encoder, action decoder and state encoder are re-initialized; all other
weights, including the language backbone, are fine-tuned. The latent world-model objective stays
enabled with its default weight of 0.1. The \texttt{front} and \texttt{overhead} views, at
$224^2$, fill the two exterior-camera slots of the pretrained model. States are normalized by
mean and standard deviation and actions by their range, with statistics computed on our
training tree. We train with AdamW (learning rate $10^{-4}$, $\beta=(0.9,0.95)$, gradient
clipping at 1.0) under a cosine schedule with 5{,}000 warmup steps, in bf16, at an effective batch
size of 32, and evaluate the step-$2{,}000$ checkpoint. The policy predicts 7 absolute poses and
executes the first 4 before replanning.

\paragraph{Closed-loop evaluation.}
Each task is evaluated on variation~0 over 20 trials, and success is the task-success signal
reported by RLBench.
MolmoAct, PAD-Depth, UWM, and VLA-JEPA run in the same rollout harness as \method{}, with scene
seeds derived from \texttt{(task, variation, trial)}, $512^2$ renders, and RLBench's
planning-based end-effector controller. An episode ends at success, at $1.5\times$ the
demonstration length, or after 20 consecutive inverse-kinematics failures; \method{} uses the
same budget but ends after 20 inverse-kinematics failures in total.
\(\pi_{0.5}\), LaWAM, Evo-1, and VLANeXt are served behind a common WebSocket interface and driven
by one model-agnostic RLBench client; their episodes end at success, after 200 executed actions,
or after 20 inverse-kinematics failures in total.

\section{Perception results under simulator replay}
\label{app:perception-results}

\tabref{tab:main} gives the per-task scores behind Figure~\ref{fig:percep_v3}, obtained with the
protocol of \appref{app:protocol} and Figure~\ref{fig:eval_pipeline}.

\begin{table}[!t]
\centering
\scriptsize
\setlength{\tabcolsep}{2.2pt}
\renewcommand{\arraystretch}{0.98}

\begin{adjustbox}{max totalsize={\textwidth}{\textheight},center}
\begin{tabular}{@{}llccccc@{}}
\toprule[1pt]
\multirow{2}{*}{\textbf{Training modalities}}
& \multirow{2}{*}{\textbf{Route}}
& \multicolumn{5}{c}{\textbf{RLBench}} \\
\cmidrule(lr){3-7}
&
& \shortstack{Close\\microwave}
& \shortstack{Toilet seat\\down}
& \shortstack{Close\\box}
& \shortstack{Meat on\\grill}
& \shortstack{Gated\\avg.}
\\
\midrule
\multicolumn{7}{@{}l}{\cellcolor{tablegray}\textbf{Future RGB} --- LPIPS $\downarrow$} \\
\menu{RGB}                       & \multirow{8}{*}{Direct} & \textbf{0.41} & 0.55 & \textbf{0.38} & 0.49 & \textbf{0.39} \\
\menu{RGB+D}                      & & 0.46 & $0.50^{\dagger}$ & 0.39 & 0.53 & 0.42 \\
\menu{RGB+S}               & & 0.45 & 0.51 & 0.39 & 0.46 & 0.42 \\
\menu{RGB+N}                    & & 0.46 & 0.50 & 0.38 & 0.54 & 0.42 \\
\menu{RGB+D+S}       & & 0.42 & \underline{0.48} & \underline{0.38} & 0.47 & \underline{0.40} \\
\menu{RGB+N+S}     & & 0.47 & $\mathbf{0.47}^{\dagger}$ & 0.39 & 0.52 & 0.43 \\
\menu{RGB+D+N}            & & 0.44 & 0.48 & 0.39 & $\underline{0.37}^{\dagger}$ & 0.42 \\
\menu{RGB+D+N+S} & & \underline{0.41} & $0.53^{\dagger}$ & 0.41 & \textbf{0.37} & 0.41 \\

\midrule
\multicolumn{7}{@{}l}{\cellcolor{tablegray}\textbf{Metric depth} --- dynamic-region AbsRel $\downarrow$ \quad \emph{external expert: Depth Anything V2}} \\
\multirow{2}{*}{\menu{RGB+D}}
& Direct                          & 0.34 & $0.57^{\dagger}$ & 0.27 & 0.38 & 0.30 \\
& External expert                & 0.44 & $0.49^{\dagger}$ & 0.42 & 0.59 & 0.43 \\
\addlinespace[1pt]
\multirow{2}{*}{\menu{RGB+D+S}}
& Direct                          & \textbf{0.27} & 0.52 & 0.39 & 0.40 & 0.33 \\
& External expert                & 0.39 & 0.49 & 0.42 & 0.57 & 0.41 \\
\addlinespace[1pt]
\multirow{2}{*}{\menu{RGB+D+N}}
& Direct                          & 0.32 & \underline{0.35} & \textbf{0.25} & \underline{0.33} & \underline{0.28} \\
& External expert                & 0.42 & 0.50 & 0.42 & $0.51^{\dagger}$ & 0.42 \\
\addlinespace[1pt]
\multirow{2}{*}{\menu{RGB+D+N+S}}
& Direct                          & \underline{0.28} & $\mathbf{0.31}^{\dagger}$ & \underline{0.25} & \textbf{0.23} & \textbf{0.26} \\
& External expert                & 0.39 & $0.53^{\dagger}$ & 0.42 & 0.49 & 0.41 \\

\midrule
\multicolumn{7}{@{}l}{\cellcolor{tablegray}\textbf{Surface normal} --- dynamic-region mean cosine $\uparrow$ \quad \emph{external expert: Lotus-G}} \\
\multirow{2}{*}{\menu{RGB+N}}
& Direct                          & \underline{0.76} & $0.68^{\dagger}$ & \underline{0.78} & $0.70^{\dagger}$ & \textbf{0.77} \\
& External expert                & 0.75 & \textbf{0.72} & \textbf{0.78} & 0.69 & 0.77 \\
\addlinespace[1pt]
\multirow{2}{*}{\menu{RGB+N+S}}
& Direct                          & 0.74 & $0.70^{\dagger}$ & 0.78 & 0.69 & 0.76 \\
& External expert                & 0.75 & $0.71^{\dagger}$ & 0.78 & 0.72 & 0.76 \\
\addlinespace[1pt]
\multirow{2}{*}{\menu{RGB+D+N}}
& Direct                          & 0.72 & 0.68 & 0.71 & 0.74 & 0.71 \\
& External expert                & 0.75 & 0.71 & 0.77 & $0.75^{\dagger}$ & 0.76 \\
\addlinespace[1pt]
\multirow{2}{*}{\menu{RGB+D+N+S}}
& Direct                          & 0.73 & 0.68 & 0.73 & \underline{0.75} & 0.73 \\
& External expert                & 0.76 & $0.70^{\dagger}$ & 0.77 & \textbf{0.77} & 0.76 \\
\midrule
\multicolumn{7}{@{}l}{\cellcolor{tablegray}\textbf{Segmentation} --- dynamic-region mIoU $\uparrow$ \quad \emph{external expert: CLIPSeg}} \\
\multirow{2}{*}{\menu{RGB+S}}
& Direct                          & \textbf{0.53} & \textbf{0.42} & \textbf{0.45} & $\mathbf{0.34}^{\dagger}$ & \textbf{0.49} \\
& External expert                & 0.19 & 0.21 & 0.28 & 0.20 & 0.23 \\
\addlinespace[1pt]
\multirow{2}{*}{\menu{RGB+D+S}}
& Direct                          & 0.42 & 0.38 & 0.42 & \underline{0.33} & 0.42 \\
& External expert                & 0.21 & 0.23 & 0.26 & 0.20 & 0.24 \\
\addlinespace[1pt]
\multirow{2}{*}{\menu{RGB+N+S}}
& Direct                          & \underline{0.49} & \underline{0.41} & \underline{0.44} & 0.30 & \underline{0.46} \\
& External expert                & 0.18 & $0.20^{\dagger}$ & 0.26 & 0.16 & 0.22 \\
\addlinespace[1pt]
\multirow{2}{*}{\menu{RGB+D+N+S}}
& Direct                          & 0.39 & 0.32 & 0.43 & 0.32 & 0.41 \\
& External expert                & 0.26 & $0.21^{\dagger}$ & 0.27 & 0.25 & 0.26 \\
\bottomrule[1pt]
\end{tabular}
\end{adjustbox}

\caption{\small
\textbf{Perception under simulator replay on four unseen RLBench tasks.}
Each task cell averages five trials at step 4{,}000 and is scored on the dynamic region against
the simulator replay induced by that route's decoded actions. For checkpoints that natively
predict a modality, adjacent rows compare direct generation with the same checkpoint's external-
expert route; the latter first predicts RGB and then applies the frozen expert named in the block
heading. Within a metric block, the gated average removes a task from
every route if any shown cell has a commanded--achieved execution gap of at least 10\,mm; those
cells remain visible and are marked $^{\dagger}$. Bold and underline mark the best and second-best
values, respectively, using the unrounded scores. Arms 0--6 share the 70/30
RLBench--ManiSkill3 training mixture. The available all-modality arm-7 checkpoint is
an RLBench-only contextual result with a different data exposure and training horizon, and is
therefore not a strictly matched ablation.
}
\label{tab:main}
\end{table}

\section{Additional qualitative results}
\label{app:qualitative}
\renewcommand{\topfraction}{0.92}
\renewcommand{\bottomfraction}{0.8}
\renewcommand{\textfraction}{0.06}
\renewcommand{\floatpagefraction}{0.6}
\setcounter{topnumber}{3}
\setcounter{totalnumber}{4}

This section collects one or two representative examples for each qualitative result of
Section~\ref{sec:experiments}. Frames are unedited simulator renders or raw generated canvases;
generated frames are outlined in green and simulator or ground-truth frames in grey. Trials are
selected by the fixed rule stated in each caption, never by inspecting the baselines' outcomes.

\subsection{Closed-loop rollouts against external policies}
\label{app:qual-closedloop}

\begin{figure}[!htbp]
  \centering
  \includegraphics[width=\linewidth]{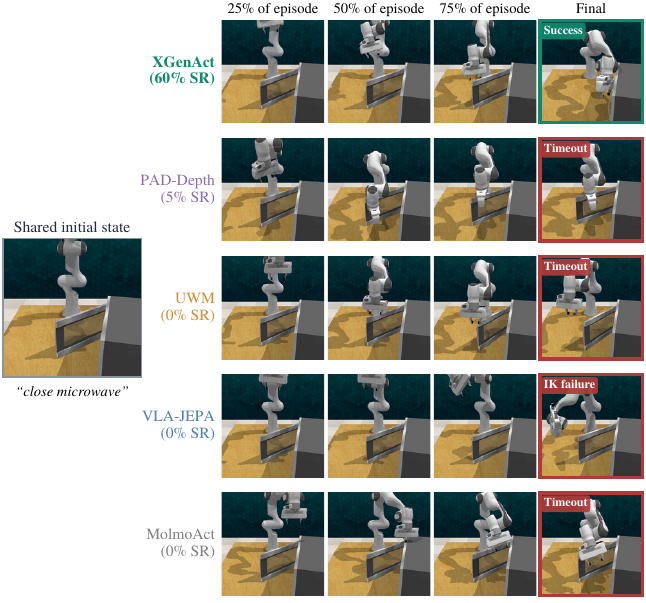}
  \caption{\small \textbf{Close microwave}, trial 0 (the first trial \method{} solves).
  \method{} closes the door; PAD-Depth, UWM and MolmoAct time out, and VLA-JEPA stops on repeated
  inverse-kinematics failures.}
  \label{fig:app-cl-microwave}
\end{figure}

\begin{figure}[!htbp]
  \centering
  \includegraphics[width=0.78\linewidth]{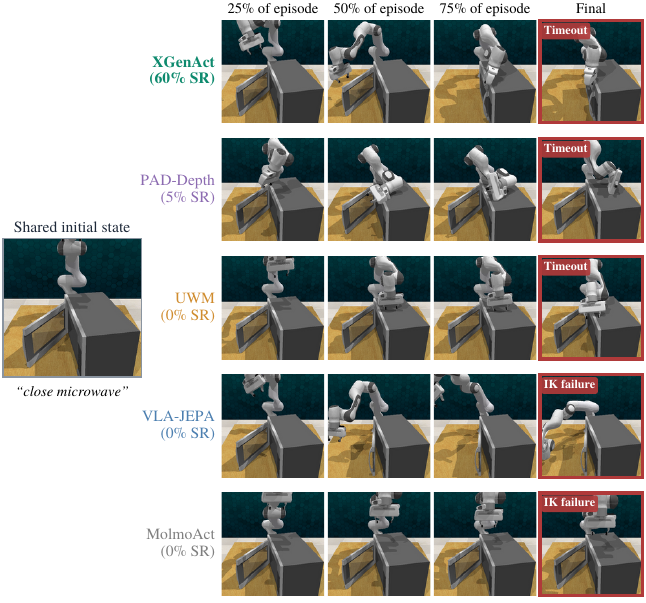}\\
  {\small (a) Close microwave, trial 3}\\[6pt]
  \includegraphics[width=0.78\linewidth]{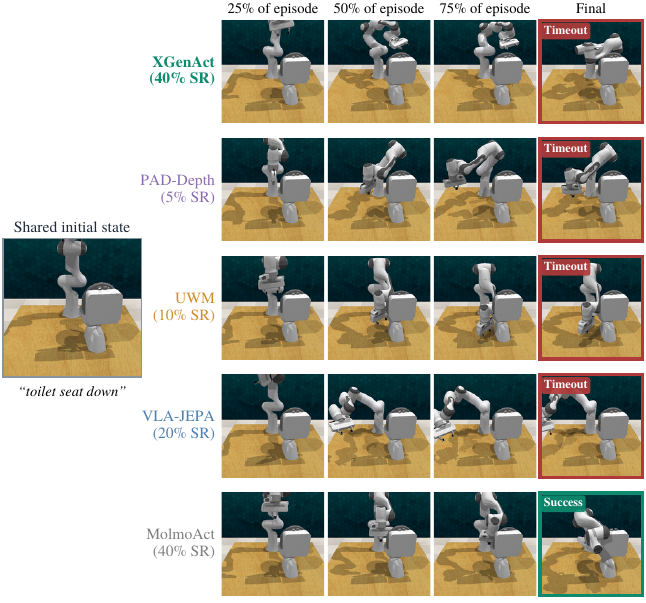}\\
  {\small (b) Toilet seat down, trial 0}
  \caption{\small \textbf{Failure cases of \method{}}: the first trial of each task that \method{}
  does not solve. Both failures are timeouts; on the same scenes every baseline also fails, except
  MolmoAct in (b).}
  \label{fig:app-cl-failures}
\end{figure}

Figures~\ref{fig:app-cl-microwave} and~\ref{fig:app-cl-failures} expand
Figure~\ref{fig:closedloop_overview} with the same runs as \tabref{tab:policy-baselines} for
\method{}, PAD-Depth, UWM, VLA-JEPA and MolmoAct; $\pi_{0.5}$, LaWAM, Evo-1 and VLANeXt run behind
a separate client (\appref{app:baselines}) whose episodes were not recorded as video. Each panel
fixes one trial, so all methods start from the same scene, and shows frames at $25/50/75\%$ of each
method's own episode and its final frame, with the outcome and termination reason. The labels give
each method's success rate over the 20 trials.

\subsection{Paired rollouts of \menu{RGB} and \method{}}
\label{app:qual-paired}

\begin{figure}[!htbp]
  \centering
  \includegraphics[width=\textwidth]{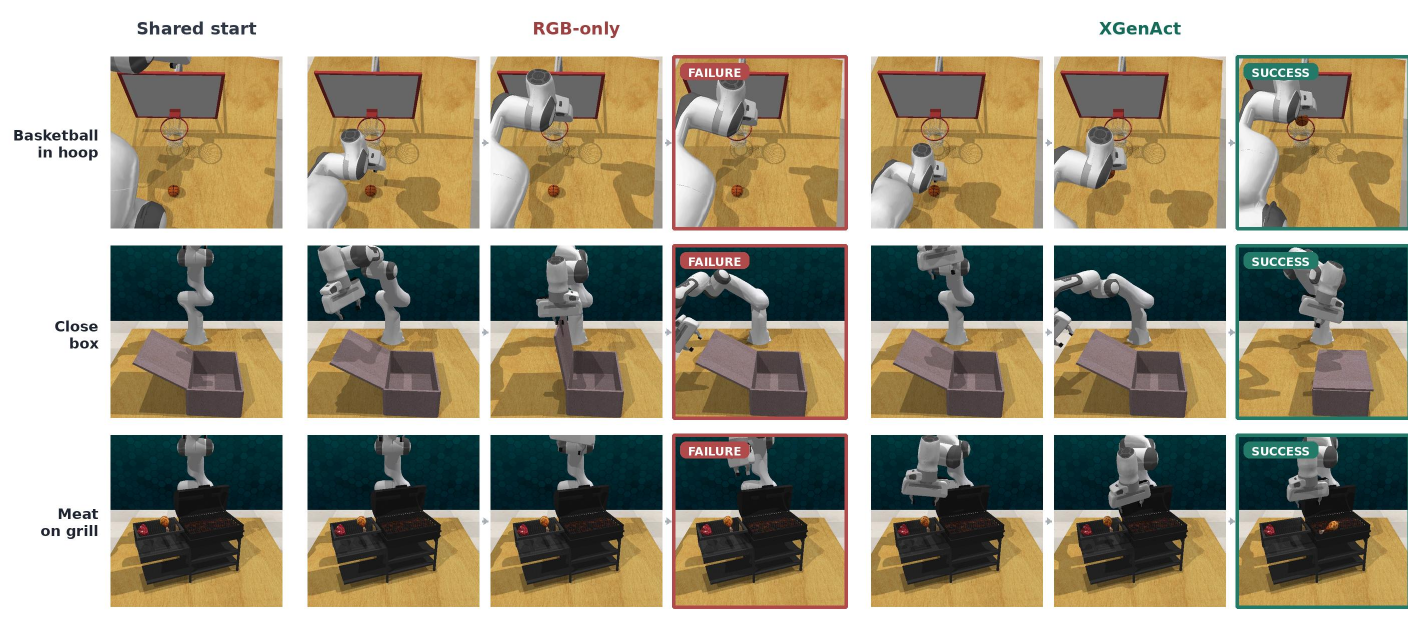}
  \caption{\small
  \textbf{Paired closed-loop executions on RLBench.}
  Each row uses the same task instance and scene seed for RGB-only and \method{}.
  The basketball row is shown from a recording-only right-shoulder camera; both policies retain
  the original front and overhead inputs used throughout the closed-loop evaluation.
  }
  \label{fig:paired_rollouts}
\end{figure}

Figure~\ref{fig:paired_rollouts} accompanies the menu ablation of Section~\ref{sec:exp_ablation}:
for three unseen tasks it shows an \menu{RGB} rollout and a \method{} rollout started from the same
task instance and scene seed.

\subsection{Imagined futures versus executed rollouts}
\label{app:qual-imagination}

\begin{figure}[!htbp]
  \centering
  \includegraphics[width=\linewidth]{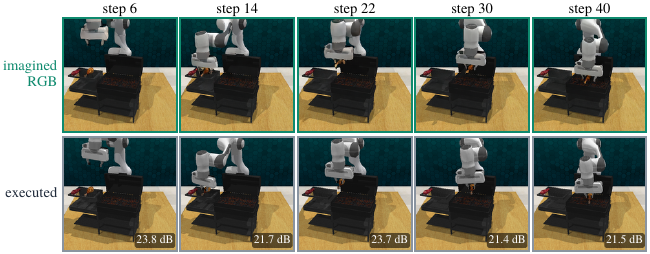}
  \caption{\small
  \textbf{Imagined versus executed future, meat on grill} (view~1, first replanning step).
  Top: RGB frames \method{} generated for step $k$. Bottom: the simulator after executing the
  decoded actions up to $k$ (corner: PSNR to the imagined frame). Across the 40-step chunk, the
  generated and executed sequences show the same grasp, transport, and placement in nearly the
  same arm poses (whole-frame PSNR $21$--$24$\,dB).}
  \label{fig:app-imagination}
\end{figure}

\begin{figure}[!htbp]
  \centering
  \includegraphics[width=0.66\linewidth]{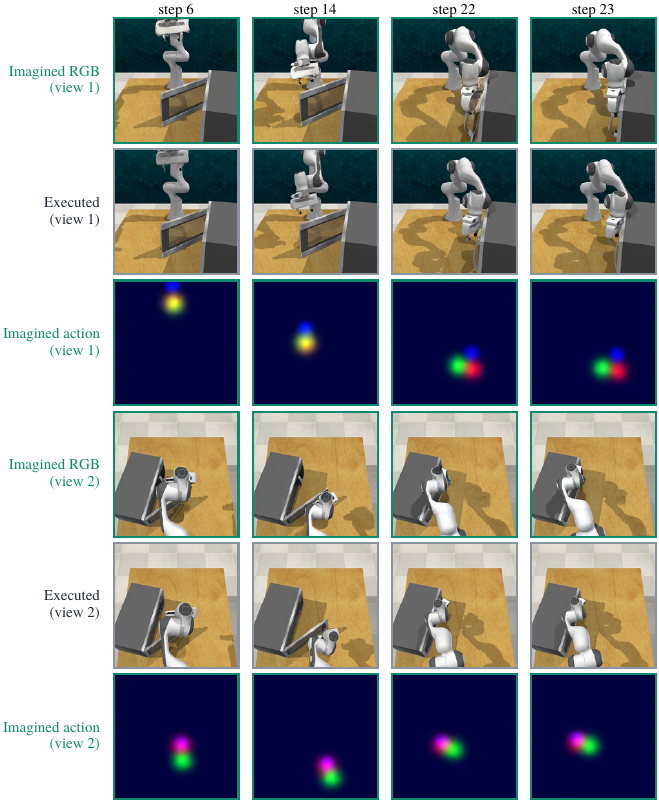}
  \caption{\small \textbf{Imagination versus execution, close microwave} (trial 0, both views). For
  each view: the generated RGB frame, the simulator after executing the decoded actions up to the
  same step, and the generated action image that produced them.}
  \label{fig:app-imag-microwave}
\end{figure}

During a closed-loop rollout, every replanning step generates a chunk of future RGB and action
images, the action images are decoded to poses, and the robot executes them. The dumps of
\appref{app:perception-results} record, for each executed step $k$, which frame of which
generated chunk it came from and what the simulator rendered after executing it, so the two can be
compared frame by frame. Figure~\ref{fig:app-imagination} shows a representative meat-on-grill
chunk, and Figure~\ref{fig:app-imag-microwave} shows both views of a close-microwave chunk. Both
use the all-modality checkpoint prompted with \ptag{video}\ptag{action} and the first chunk of the
lowest-index successful trial; the latter omits the first four steps, which the rollout replaces by
a ramp from the current pose.

\subsection{Open-loop generation across tasks}
\label{app:qual-generation}

\begin{figure}[!htbp]
  \centering
  \includegraphics[width=\linewidth,height=0.7\textheight,keepaspectratio]{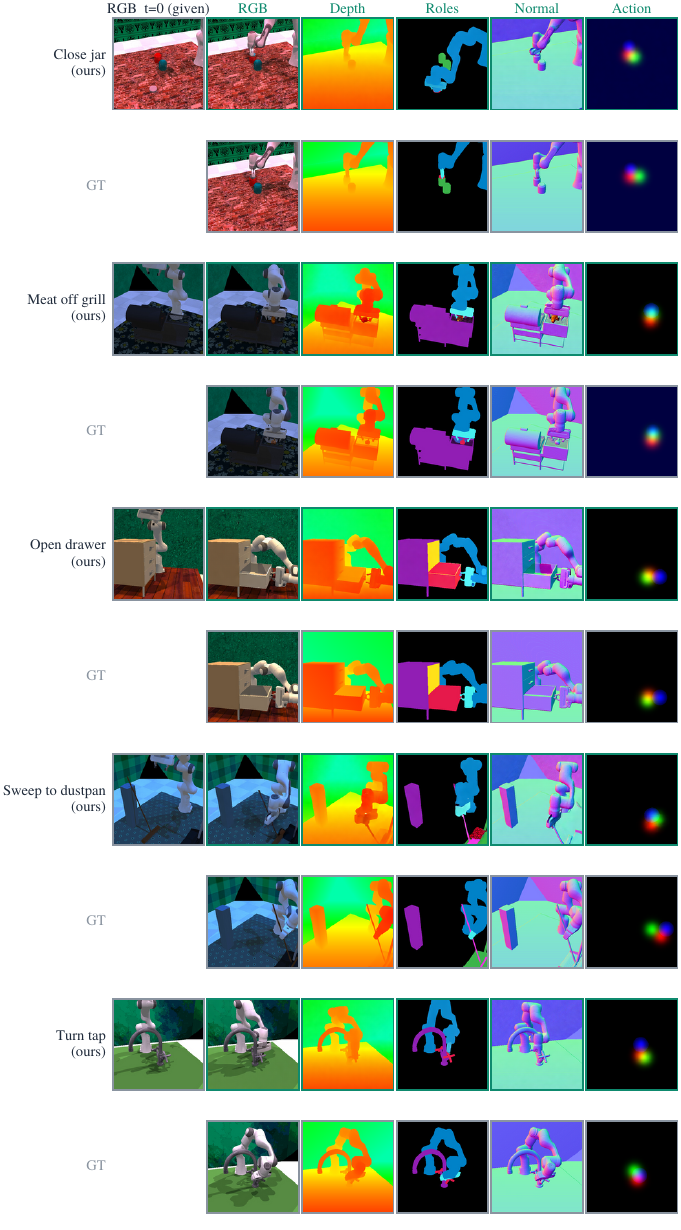}
  \caption{\small \textbf{Open-loop generation across output spaces.} For each episode, the upper
  row is generated (frame 40 of 40, view 1) and the lower row is the demonstration. Each output
  space is a separate prompt of the same checkpoint, so the arm pose can differ slightly between
  columns.}
  \label{fig:app-grid-seen}
\end{figure}

\begin{figure}[!htbp]
  \centering
  \includegraphics[width=\linewidth]{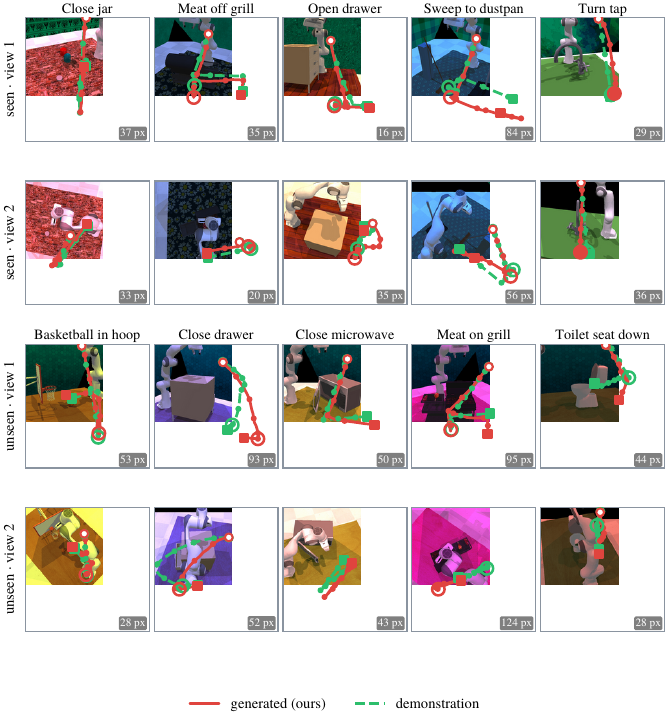}
  \caption{\small \textbf{End-effector paths decoded from generated action images} (red) versus the
  demonstration (green, dashed), drawn on the frame-0 observation of both views; the top two rows
  are training tasks and the bottom two held-out tasks. Each point is the centroid of the position
  blob of one frame; open circle: start, square: end, ring: first frame with the gripper closed. The
  corner value is the mean 2-D distance between the two paths in pixels ($512^2$ images); it is
  not the 3-D pose error of the tables.}
  \label{fig:app-traj}
\end{figure}

The last group is open loop. For demonstration episodes of training and held-out tasks, the
all-modality checkpoint trained on the $70/30$ RLBench--ManiSkill3 mixture (step 4{,}000) receives
only the frame-0 observation and the current end-effector pose of two views. With plan
\texttt{IIII} it generates 40 future frames, once per template (50 flow steps, guidance 7.5,
seed 42); the ground truth is the demonstration. Figure~\ref{fig:app-grid-seen} shows the last
generated frame in every output space, and Figure~\ref{fig:app-traj} decodes the generated action
images into end-effector paths on training and held-out tasks.

\end{document}